\documentclass[11pt]{article}

\usepackage[preprint]{acl}

\usepackage{times}
\usepackage{latexsym}
\usepackage{booktabs}
\usepackage{tcolorbox}
\usepackage{xcolor}

\usepackage[T1]{fontenc}

\usepackage[utf8]{inputenc}

\usepackage{microtype}
\usepackage{amsmath}
\usepackage{amssymb}

\usepackage{algorithm}
\usepackage{algorithmic}
\usepackage{amsmath}
\usepackage{amssymb}
\usepackage{inconsolata}

\usepackage{graphicx}

\newtheorem{theorem}{Theorem}
\title{Learning to Adapt SFT Data for Better Reasoning Generalization}

\author{
 \textbf{Lisong Sun\textsuperscript{1}\thanks{Equal contribution.}},
 \textbf{Li Wang\textsuperscript{1}\footnotemark[1]},
 \textbf{Chen Zhang\textsuperscript{1}},
 \textbf{Jinyang Wu\textsuperscript{2}},
\\
\textbf{Kui  Zhang\textsuperscript{1}},
 \textbf{Tianhao Peng\textsuperscript{3}\thanks{Corresponding authors: Tianhao Peng: \href{mailto:tianhao.peng@ntu.edu.sg}{tianhao.peng@ntu.edu.sg}, Wenjun Wu: \href{mailto:wwj09315@buaa.edu.cn}{wwj09315@buaa.edu.cn}.}},
 \textbf{Wenjun Wu\textsuperscript{1,4,5}\footnotemark[2]}
\\
 \textsuperscript{1}Beihang University,
 \textsuperscript{2}Tsinghua University,
 \textsuperscript{3}Nanyang Technological University
\\
 \textsuperscript{4}Hangzhou International Innovation Institute
\\
 \textsuperscript{5}Beijing Advanced Innovation Center for Future Blockchain and Privacy Computing
}

\begin{document}
\maketitle
\begin{abstract}
Large language models (LLMs) have achieved remarkable progress, with post-training playing a crucial role in enhancing their reasoning capabilities. Among post-training paradigms, supervised fine-tuning (SFT) is widely used: it leverages external data to provide dense supervision and enables efficient training. However, directly fine-tuning on expert data can hurt generalization when the data distribution is mismatched with the target model's own distribution. In this work, we propose Data Adaptation for Reasoning Tuning (DART), which formulates the use of a fixed, potentially distributionally misaligned SFT dataset as an optimization problem over demonstration transformations. DART trains a mapper model with reinforcement learning to convert original SFT data into model-adapted supervision that better matches the target model's distribution and learning preferences. The transformed data are then used for SFT, allowing the target model to better exploit external supervision. Experiments across multiple models and datasets show that DART improves generalization, achieves higher training efficiency than direct RL, and helps models surpass standard SFT. Our code is available at \url{https://anonymous.4open.science/r/DART525E50D}.
\end{abstract}

\section{Introduction}
Large language models (LLMs)~\cite{yang2025qwen3,team2026kimi,guo2025deepseek} have achieved remarkable progress, and post-training has further strengthened their capabilities in mathematical reasoning~\cite{guan2025rstar,chen2026kg}, code generation~\cite{yang2025code}, tool use~\cite{jin2025search,xue2025simpletir}, and other complex reasoning tasks~\cite{feng2026group,wang2026tamtrl}. Among post-training paradigms, supervised fine-tuning (SFT) remains widely adopted due to its simplicity and scalability. However, SFT is highly sensitive to the quality and diversity of fine-tuning data. Low-quality~\cite{chen2024alpagasus} or poorly matched supervision, such as flawed reasoning traces, samples concentrated in narrow domains, or examples that are misaligned with the target model's distribution, can provide ineffective or even misleading learning signals, thereby reducing training efficiency and degrading downstream performance~\cite{huang2026fine,havrilla2024understanding}. These challenges make data quality central bottlenecks in SFT.

\begin{figure}[t]
    \centering
    \includegraphics[width=\linewidth]{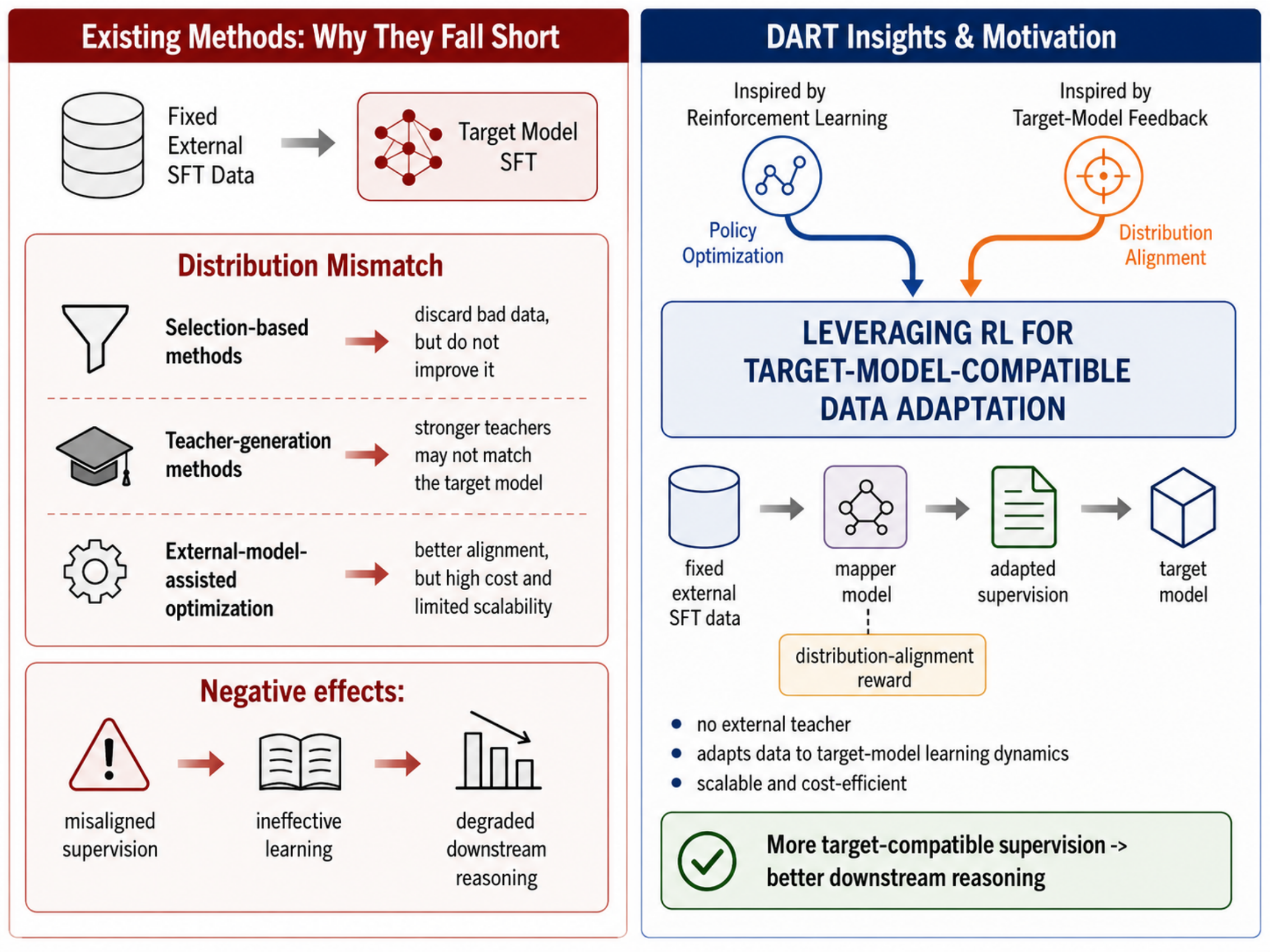}
    \caption{Comparison between existing SFT data optimization methods and our approach.}
    \label{fig:motivation}
\end{figure}

As shown in Figure~\ref{fig:motivation}, to address the bottleneck of SFT data quality, existing studies have explored several data-centric strategies. Selection-based methods filter training examples using human annotation, LLM-based assessment, or heuristic criteria such as diversity and difficulty~\cite{li2024superfiltering,zhang2025d3}. However, they mainly discard undesirable examples and often rely on carefully engineered selection rules. Another line of work employs stronger external models to generate higher-quality supervision~\cite{lee2024llm2llm}. Yet stronger teachers do not necessarily yield better students~\cite{xu2025stronger,li2025small}, since externally generated demonstrations can deviate from the target model's behavior distribution and may impair generalization~\cite{li2025scar,yang2024self,shenfeld2025rl,huang2026fine}. Recent methods alleviate this mismatch by using external models to optimize existing data while involving the student through selection~\cite{li2024selective,zhang2026best} or alternating generation~\cite{huang2026fine}. Nevertheless, their dependence on external models incurs substantial generation and refinement costs, making them difficult to scale. In practice, leveraging external SFT data is efficient, as large amounts of high-quality supervision data are available from diverse sources. However, these data are often distributionally misaligned with the target model due to differences in model behaviors, domains, or response styles, which can limit the effectiveness of direct SFT. This motivates our central research question:

\emph{Given only a fixed external SFT dataset that is distributionally misaligned with the target model, can we enable the model to self-optimize its supervision by adapting the given data to better match its own learning needs, without relying on any external model?}

To answer this question, we make the following technical contributions:

\begin{itemize}
    \item \textbf{Problem Identification and Formulation.}
    We identify a practical yet underexplored challenge in SFT data optimization: how to effectively exploit fixed external SFT data that may be distributionally misaligned with the target model through self-optimization. To study this challenge, we introduce a self-optimization setting for fixed-data adaptation and formulate it as an explicit optimization problem, where the objective is to adapt the given supervision into a more target-model-compatible form that improves downstream reasoning performance after fine-tuning.

    \item \textbf{Data Adaptation for Reasoning Tuning.} We introduce Data Adaptation for Reasoning Tuning (DART), an RL-based method for optimizing SFT data. DART first trains a mapper model with reinforcement learning to transform the fixed SFT dataset, and we design a distribution-alignment reward to encourage the adapted data to better match the learning dynamics of the target model. The trained mapper is then used to generate an adapted SFT dataset, on which the target model is fine-tuned to improve reasoning performance.

    \item \textbf{Empirical Effectiveness.} We conduct extensive experiments across multiple models and datasets. The results demonstrate that DART consistently improves the effectiveness of SFT and achieves better performance than directly applying RL to the target model, highlighting the benefit of adapting external supervision before fine-tuning.
\end{itemize}

\section{Related work}
\subsection{LLM Post-training}
RL and SFT are two widely adopted paradigms for post-training LLMs. While RLVR has been shown to better preserve generalization, it often suffers from lower training efficiency~\cite{wachi2026relative,nguyen2026adaptive} and may struggle to push models beyond their existing capability frontier~\cite{chen2026does,wu2025invisible}. In contrast, SFT offers a simple and efficient way to instill desired reasoning patterns through direct supervision from expert demonstrations. However, stronger teachers do not necessarily produce better students~\cite{xu2025stronger,li2025small}: when the demonstrated reasoning deviates from the student model's own distribution, the imitation objective can induce distribution mismatch and impair reasoning generalization~\cite{li2025scar,yang2024self,shenfeld2025rl,huang2026fine}. Unlike prior studies, we investigate a complementary direction: optimizing a fixed SFT dataset to better align with the student model's distribution, thereby improving the generalization of fine-tuning.

\subsection{SFT Data Optimization}
Recent studies have increasingly recognized the quality of fine-tuning data as a critical determinant of downstream post-training effectiveness~\cite{wei2022chain,luong2024reft,feng2025cot,guha2025openthoughts,li2025small}. To improve post-training performance, existing work has explored data optimization from multiple perspectives, including enhancing data quality and diversity~\cite{zhou2023lima,chen2024alpagasus,xia2024less}, as well as improving reasoning traces through multi-path sampling, filtering, and evolutionary refinement~\cite{feng2025cot,guha2025openthoughts,ahmad2025opencodereasoning}. These approaches generally aim to construct higher-quality or more informative supervision signals, but they often treat the fine-tuning dataset as model-agnostic and do not explicitly account for its distributional compatibility with the target model. More closely related to our work, TESSY performs distribution alignment by synthesizing student-consistent data, assigning capability-related tokens to a stronger teacher while delegating style-related tokens to the student~\cite{huang2026fine}. In contrast to these prior approaches, our method directly optimizes a given fine-tuning dataset under an explicit distribution-alignment objective, without relying on external models, thereby adapting the supervision data to the target model and improving the generalization of SFT.

\section{Self-Optimization for SFT Data Adaptation}
\label{sec:question define}
We consider the problem of improving reasoning fine-tuning with a fixed external SFT dataset. Formally, let $\mathcal{D}=\{(x_i,y_i)\}_{i=1}^{n}$ denote the original SFT dataset, where $x_i$ is the input prompt and $y_i$ is the provided demonstration. Instead of directly fine-tuning the target model on $\mathcal{D}$, we aim to self-optimize the given supervision by learning a transformation function $f_{\phi}$ that adapts each original demonstration into a form better aligned with the target model, i.e., $\tilde{y}_i=f_{\phi}(x_i,y_i)$. This yields a transformed dataset $\tilde{\mathcal{D}}_{\phi}=\{(x_i,\tilde{y}_i)\}_{i=1}^{n}$, which is then used for SFT. The objective is to find a transformation function that enables the target model to achieve stronger downstream reasoning performance after fine-tuning on the transformed dataset:
\begin{equation}
\begin{array}{ll}
\displaystyle \max_{\phi} 
& \mathrm{Perf}\big(\pi_{\theta^{\star}(\phi)}\big) \\[3pt]
\mathrm{s.t.}
& \displaystyle
\theta^{\star}(\phi)
=
\arg\max_{\theta}
\frac{1}{n}
\sum_{i=1}^{n}
\log \pi_{\theta}(\tilde{y}_i \mid x_i), \\[8pt]
& \tilde{y}_i=f_{\phi}(x_i,y_i), \quad i\in[n].
\end{array}
\label{eq:data_adaptation_obj}
\end{equation}

\noindent Here, $\mathrm{Perf}(\cdot)$ denotes the evaluation performance on downstream reasoning tasks. Formally, we establish Theorem~\ref{thm:theorem} to characterize the performance gain achieved through SFT data optimization.

\begin{theorem}
\label{thm:theorem}
Let $p_0(y\mid z)$ and $p_\phi(y\mid z)$ denote the Original-SFT and the optimized data distributions for a reasoning pattern $z$, respectively. Let $\mathcal{C}_z$ be the set of correct reasoning trajectories and $\mathcal{M}_z$ be the base-compatible region induced by the base model distribution. Define the effective SFT supervision probability as
\[
\rho_p(z)=P_p(\mathcal{C}_z\cap\mathcal{M}_z\mid z).
\]
For Original-SFT, let
\[
\begin{aligned}
c_0(z) &= P_{p_0}(\mathcal{C}_z\mid z), \\
m_0(z) &= P_{p_0}(\mathcal{M}_z\mid\mathcal{C}_z,z).
\end{aligned}
\]
Since the optimized data have passed correctness verification and improve distributional compatibility, we have
\[
\begin{aligned}
P_{p_\phi}(\mathcal{C}_z\mid z) &= 1, \\
P_{p_\phi}(\mathcal{M}_z\mid\mathcal{C}_z,z) &\ge m_0(z)+\Delta_m(z).
\end{aligned}
\]
where $\Delta_m(z)>0$ and $m_0(z)+\Delta_m(z)\le 1$. If the model can reliably acquire pattern $z$ after observing at least $k_z$ effective samples among $n_z$ training samples, then
\[
\mathrm{Acc}_{p_\phi}(z)>\mathrm{Acc}_{p_0}(z).
\]
Consequently, over the test distribution of reasoning patterns,
\[
\mathrm{Acc}_{\mathrm{Optimized}}>
\mathrm{Acc}_{\mathrm{Original\text{-}SFT}}.
\]
\end{theorem}

\noindent \textbf{Remark:} The proof is provided in the Appendix \ref{appendix:proof}. Theorem \ref{thm:theorem} offers useful insights into the role of data distribution alignment. Original-SFT may contain correct reasoning trajectories, but many of them may fall outside the compatible region of the base model's data distribution, making them less effective for the target model. Our method increases the conditional probability that correct trajectories lie within the base model's compatible distributional region. As a result, the probability of observing sufficient effective supervision increases, which enhances the model's performance.

\section{Method}
\begin{figure*}[!t]
    \centering
    \includegraphics[width=1\textwidth]{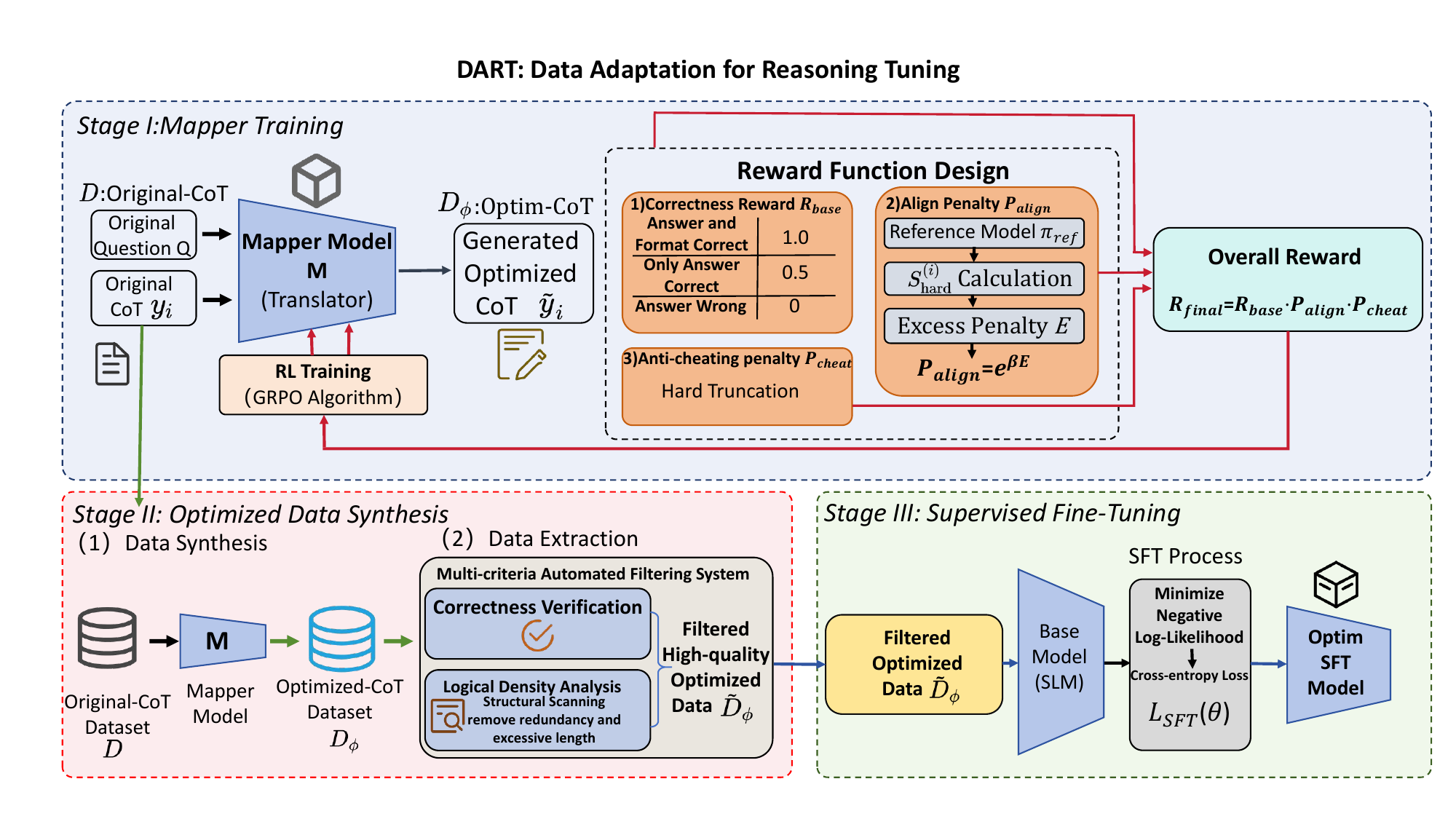}
    \caption{Framework of DART Method. DART consists of (1) Mapper Training, a mapper model is trained via RL to rephrase original CoTs and achieve distribution alignment, (2) Optimized Data Synthesis, the trained mapper model generates optimized CoTs, which are then filtered to retain high-quality samples, and (3) Supervised Fine-Tuning, the target model is fine-tuned on the optimized dataset to enhance reasoning performance.}
    \label{fig:method}
\end{figure*}
We propose an RL based framework that trains a mapper model to learn how to generate reasoning steps conforming to the model's data distribution, thereby aligning with the target model's data distribution and optimizing the fine-tuning data for the target model. The goal is to improve the performance of the target model via SFT with optimized CoT data. The mapper model and the target model use the same pretrained base model. Rephrasing the original CoT from the mapper model's own perspective brings the optimized CoT closer to the target model's data distribution, without relying on any external model. The framework consists of the following three phases: (1) \textbf{Mapper Training}: A mapper model is trained using an RL algorithm to transform original CoTs $y_i$ into high-quality CoTs $\tilde{y_i}$ that better conforms to the target model's data distribution. (2) \textbf{Optimized Data Synthesis}: The mapper model is employed to generate optimized data, which are then subjected to automated verification of both correctness and logical density, thereby extracting high-quality data for SFT; (3) \textbf{Supervised Fine-Tuning}: The base model is fine-tuned using the optimized data to improve performance in a better aligned distribution. The complete pseudocode is provided in Appendix \ref{pseudocode}.

\subsection*{Stage I: Mapper Training}
Directly solving the optimization problem in section \ref{sec:question define} is highly challenging, as the gradient of the variable $\phi$ must propagate through the entire SFT trajectory, which often leads to severe gradient vanishing and computational infeasibility under discrete text generation and large-scale parameter spaces. In contrast, RL can model discrete text generation as a sequential decision-making problem, directly guiding policy updates via reward signals.
So this study formulates the data optimization process as an RL task. The mapper model \( M \) aims to transform the original reasoning paths $y_i$ into a high-quality data distribution $\tilde{y_i}$ that is more compatible with the cognitive features of target model. During the training process, the mapper model \( M \) is trained to align the generated reasoning paths with the target model's data distribution, while preserving correctness, eliminating redundancy, and strengthening core logical steps.

\subsubsection*{Reward Function Design}
The goal of reinforcement learning in our framework is to optimize the mapper model for CoT data transformation. The learned mapper is expected to generate transformed CoT data that better aligns with the distribution of the target model, while maintaining the fidelity, coherence, and reasoning quality of the original CoT data. Accordingly, we design the following reward function: 1) To prevent the model from deviating from the natural language distribution, a penalty term for distribution alignment based on the reference model \( \pi_{\text{ref}} \) is introduced. Let the generated sequence for the \(i\)-th sample be \( y_{i} = (y_{i1}, y_{i2}, \dots, y_{iN}) \).  For each token $y_{it}$, the generation probability under the reference model, $\pi_{\text{ref}}(y_{it} \mid y_{i,<t})$, is computed. A probability threshold $\tau$ is set. When $\pi_{\text{ref}}(y_{it} \mid y_{i,<t}) < \tau$, the token is identified as a low-probability token deviating from the base model's data distribution. The cumulative deviation penalty sum \( S_i^{\text{hard}} \) is then calculated. 
\begin{equation}
\begin{aligned}
S_i^{\text{hard}}
&= \sum_{t=1}^{N}
\mathbb{I}\!\left(
\pi_{\text{ref}}(y_{it} \mid y_{i,<t}) < \tau
\right) \\
&\quad \cdot
\log \pi_{\text{ref}}(y_{it} \mid y_{i,<t}).
\end{aligned}
\label{eq:s_hard}
\end{equation}
To prevent the penalty from being too severe, a tolerance \( T \) and a decay coefficient \( \beta \) are introduced to define the excess penalty value \( E \). 
\begin{equation}
E = \min(0, S_i^{\text{hard}} - T)
\label{eq:excess_penalty}
\end{equation}
The final penalty factor is given by \( P_{\text{align}} = e^{\beta \cdot E} \). 
2) To ensure the correctness of the model's generated results, a reward score \( R_{\text{base}} \) is assigned based on answer format and outcome correctness.
\begin{equation}
R_{\text{base}}(y) =
\begin{cases} 
1.0, & \text{correct and standard format} \\
0.5, & \text{correct but non-standard format} \\
0.0, & \text{incorrect or no answer}
\end{cases}
\label{eq:r_base}
\end{equation}
3) To prevent the model from exploiting shortcut behaviors to obtain high rewards without producing meaningful reasoning, we introduce an anti-cheating penalty \(P_{\text{cheat}}\). It is computed based on a set of rule-based indicators. The detailed rules and corresponding formulas are provided in the Appendix \ref{appendix:imple_details}.

Integrating the above components, for any complete generated sequence, the overall reward is computed as follows:
\begin{equation}
R_{\text{final}} = R_{\text{base}} \cdot P_{\text{align}} \cdot P_{\text{cheat}}
\label{eq:reward_final}
\end{equation}
Through the data optimization strategy, we produce training samples that better conform to the the target model's data distribution while exhibiting greater logical rigor and learning efficiency, thus establishing the data foundation for mitigating the generalization degradation induced by SFT.

\subsection*{Stage II: Optimized Data Synthesis}
After the mapper model \( M \) is trained, we use it as the core inference engine to perform large-scale logical restructuring and reasoning transformation on the original training corpus 
\(\mathcal{D}=\{(x_i,y_i)\}_{i=1}^{n}\).
(1) Data Synthesis. In the synthesis stage, we use the trained mapper model $M$ to translate and optimize the original CoT dataset $\mathcal{D}$. For each sample $(x_i, y_i) \in \mathcal{D}$, the mapper generates an optimized CoT $\tilde{y}_i$, yielding the transformed sample $(x_i, \tilde{y}_i)$:
\begin{equation}
\tilde{y}_i = M(x_i, y_i), \qquad (x_i, y_i) \xrightarrow{M} (x_i, \tilde{y}_i).
\end{equation} 
This operation transforms the implicit reasoning capabilities acquired during the RL stage into explicit, high-quality CoT samples, ensuring alignment with the actual data distribution while reducing redundancy, thereby yielding an optimized dataset \(\mathcal{D}_{\phi}\).
(2) Data Extraction: We adopt a multi-criteria automated filtering system \( F \) to select high-quality samples via correctness verification and logical density analysis, where the final retained dataset is defined as \(\tilde{\mathcal{D}}_{\phi}=\{(x_i,\tilde{y}_i)\mid F(x_i,\tilde{y}_i)=1\}_{i=1}^{n}\). Dual verification ensures answer-ground truth consistency, while structural scanning removes redundant and overly long samples. This enhances logical density and reduces distributional noise, mitigating performance degradation in the target model during SFT caused by unstable data quality.
\subsection*{Stage III: Supervised Fine-Tuning}
The target model is fine-tuned using the optimized dataset $\tilde{\mathcal{D_\phi}}$ to systematically validate the substantial improvements brought by the RL guided data optimization strategy through data distribution alignment. SFT typically adopts the cross-entropy loss function as its core objective. Given the optimized training dataset $\tilde{\mathcal{D}}_\phi = \{(x_{i}, \tilde{y}_{i})\}_{i=1}^{n}$, where $x_{i}$ is the input sequence and $\tilde{y}_{i}$ is the optimized target output sequence, the goal is to minimize the negative log-likelihood:
\begin{equation}
\mathcal{L}_{\text{SFT}}(\theta) = -\frac{1}{n} \sum_{i=1}^{n} \sum_{t=1}^{|\tilde{y}_{i}|} \log P_{\theta}(\tilde{y}_{it} \mid x_{i}, \tilde{y}_{i,<t})
\label{eq:sft_loss}
\end{equation}

\noindent During the SFT process, all training samples are uniformly augmented with structured CoT labels. The significance of this treatment lies in its ability to ensure that the observed performance improvements stem solely from substantive enhancements in the quality of the reasoning paths, rather than being confounded by specific formatting instructions.

\section{Experiments}
\subsection{Experiment Setup}
\paragraph{Datasets} To ensure comprehensiveness in evaluation, this study constructs a multi-dimensional experimental corpus. For the training dataset, we use MATH 3 to 5 \cite{DBLP:journals/corr/abs-2103-03874}. The evaluation datasets are divided into the following categories: (1) Mathematical Reasoning: SVAMP \cite{DBLP:journals/corr/abs-2103-07191} is used to assess robustness in basic mathematical operations. (2) Logical Reasoning: Boolean Expressions \cite{suzgun2022challengingbigbenchtaskschainofthought} is used to examine pure symbolic logical calculus, Web of Lies \cite{suzgun2022challengingbigbenchtaskschainofthought} is used to test the model's state-tracking ability under multi-level nested statements, and ProntoQA \cite{saparov2023languagemodelsgreedyreasoners} is used to evaluate the model's capability in formal reasoning and adherence to logical rules. (3) Code Capability:  MBPP \cite{austin2021programsynthesislargelanguage} and Human-Eval \cite{chen2021evaluatinglargelanguagemodels} are employed to evaluate the model's algorithmic reasoning performance. In addition, ARC-Challenge \cite{clark2018thinksolvedquestionanswering} is selected as the target dataset for transfer experiment. Among these, both Boolean Expressions \cite{suzgun2022challengingbigbenchtaskschainofthought} and Web of Lies \cite{suzgun2022challengingbigbenchtaskschainofthought} are high-quality subsets of Big Bench Hard \cite{suzgun2022challengingbigbenchtaskschainofthought}. We provide detailed statistics of all datasets in Appendix~\ref{appendix:dataset_details}.

\paragraph{Baselines} To investigate performance differences across optimization pathways, we compare our method with several representative baselines: GRPO~\cite{shao2024deepseekmath}, a reinforcement-learning-based optimization method; Original-SFT, which fine-tunes the target model on the original CoT dataset; TESSY~\cite{huang2026fine}, which alternates between student- and teacher-generated CoTs to align data distributions before fine-tuning; and STaR~\cite{zelikman2024star}, which iteratively generates CoTs with the model itself and fine-tunes on filtered correct trajectories.

\paragraph{Implementation} We use Qwen2.5-0.5B-Instruct and Qwen3-0.6B~\cite{yang2025qwen3} as the backbone models. During the RL training stage of the mapper model, the learning rate is set to $\alpha = 1 \times 10^{-6}$, and the maximum response length of the model is 4096 tokens. During SFT, we inject structural tags for format alignment, ensuring consistent initial features between the original and optimized data. During the experiments, stage I runs for 1 epoch and SFT runs for 3 epochs. All experiments are carried out on 2 A100 GPUs with 40GB memory. Further implementation details are provided in Appendix~\ref{appendix:imple_details}.

\paragraph{Evaluation} We use answer accuracy as the primary evaluation metric, which directly measures whether model answer matches the ground-truth solution. To obtain a more robust estimate under stochastic decoding, we adopt the avg@4 evaluation protocol. During the evaluation phase, decoding is performed with the temperature set to 0.7, and vLLM is adopted to accelerate inference.

\subsection{Main Results}
As shown in Table~\ref{tab:qwen3-main-acc}, our method achieves the best average accuracy of 58.42\% on Qwen3-0.6B~\cite{yang2025qwen3}, consistently outperforming all baseline methods. Compared to the Base model, the overall performance improves by 2.80\%; compared to GRPO, it further outperforms by 2.4\%; compared to STaR, it improves by 1.94\%; and compared to TESSY, it improves by 3.74\%. Original-SFT suffers from a severe decline in generalization ability due to the distribution mismatch between the original CoT and the target model's data distribution, with average performance dropping by 9.55\% relative to Base. Our method successfully overcomes this issue and enhances the model's practical performance.
On Qwen2.5-0.5B-Instruct\cite{qwen2025qwen25technicalreport}, our method surpasses the base model, GRPO, STaR, and TESSY by 3.84\%, 2.56\%, 3.57\%, and 9.43\%, respectively. These experimental results validate the advantages of the RL based data optimization strategy in improving both model reasoning performance and generalization ability by addressing the data distribution mismatch problem. Models trained on the RL-optimized dataset outperform both the baseline models across multiple comprehensive capability tests, demonstrating the substantial potential of RL in data augmentation and quality optimization.

\begin{table*}[!ht]
\centering
\caption{Performance (\%) of Qwen3-0.6B and Qwen2.5-0.5B-Instruct models across representative reasoning benchmarks under various methods. The \textbf{bold} and \underline{underline} indicate the best and second-best results, respectively.}
\small
\renewcommand{\arraystretch}{1.1}
\setlength{\tabcolsep}{3pt}
\begin{tabular*}{\textwidth}{@{\extracolsep{\fill}}l|c|ccc|cc|c}
\toprule
& \multicolumn{1}{c|}{\textbf{Math}} & \multicolumn{3}{c|}{\textbf{Logical Reasoning}} & \multicolumn{2}{c|}{\textbf{Code}} 
& \\

\textbf{Method} & \textbf{SVAMP} & \textbf{Boolean Expressions} & \textbf{Web of Lies} & \textbf{ProntoQA} & \textbf{MBPP} & \textbf{Human-Eval} & \textbf{Average} \\
\midrule
\multicolumn{8}{c}{\textit{Models based on Qwen3-0.6B}} \\
\midrule

Base & 79.03 & 72.60 & 50.50 & 59.95 & 34.01 & \underline{37.65} & 55.62 \\
Original-SFT & \underline{80.90} & 63.00 & 20.20 & 49.95 & 30.52 & 31.86 & 46.07 \\
GRPO & 80.00 & \underline{73.70} & 49.50 & 60.65 & 34.52 & 37.50 & 55.98 \\
TESSY & 79.00 & 72.80 & 52.40 & 59.85 & 31.11 & 32.93 & 54.68 \\
STaR & 79.70 & 73.00 & \underline{53.40} & \underline{61.00} & \underline{34.60} & 37.20 & \underline{56.48} \\
\textbf{DART (ours)} & \textbf{81.37} & \textbf{73.90} & \textbf{59.20} & \textbf{63.05} & \textbf{34.91} & \textbf{38.11} & \textbf{58.42} \\

\midrule
\multicolumn{8}{c}{\textit{Models based on Qwen2.5-0.5B-Instruct}} \\
\midrule

Base & 53.95 & 46.60 & \underline{41.10} & \underline{44.80} & 29.72 & 30.34 & 41.09 \\
Original-SFT & 56.17 & \textbf{56.20} & 29.80 & 36.70 & 19.75 & 25.61 & 37.37 \\
GRPO & \underline{56.40} & \underline{50.30} & 39.70 & \textbf{46.00} & \underline{29.98} & \underline{31.86} & \underline{42.37} \\
TESSY & 47.80 & 39.40 & 25.80 & 39.50 & 28.88 & 31.62 & 35.50 \\
STaR & 55.20 & 47.10 & 40.40 & \underline{44.80} & 29.88 & 30.79 & 41.36 \\
\textbf{DART (ours)} & \textbf{61.45} & 50.10 & \textbf{46.90} & 44.05 & \textbf{34.01} & \textbf{33.08} & \textbf{44.93} \\

\bottomrule
\end{tabular*}
\label{tab:qwen3-main-acc}
\end{table*}

\subsection{Training Dynamics}
\begin{figure}[!t]
    \centering
    \includegraphics[width=\columnwidth]{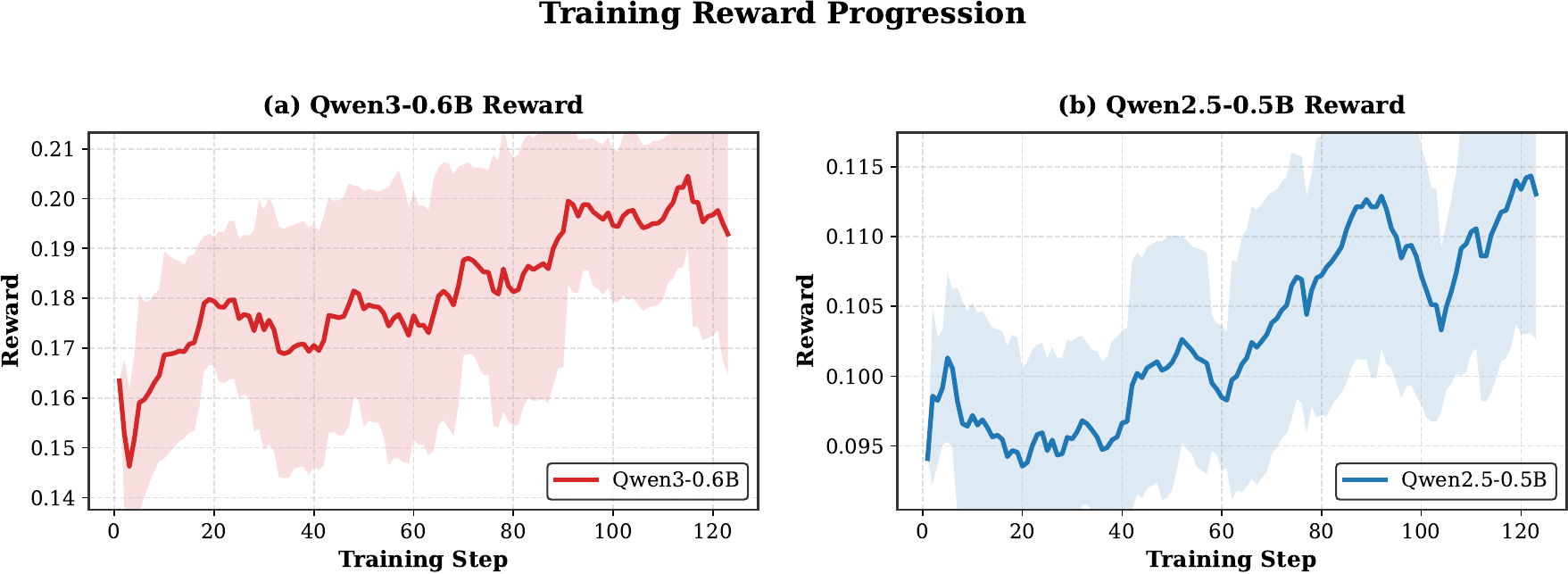}
    \caption{Training reward progression of DART on (a) Qwen3-0.6B and (b) Qwen2.5-0.5B-Instruct.}
    \label{fig:reward_progression}
\end{figure}

We further provide the training dynamics of DART to illustrate its optimization behavior. During the RL training phase, the reward steadily increases and gradually converges, successfully yielding the desired mapper model, as shown in Figure~\ref{fig:reward_progression}. The consistent upward trend of both reward curves demonstrates that the mapper model progressively improves its ability to generate reasoning paths aligned with the target distribution, validating the effectiveness of our reward design in guiding policy updates toward the target distribution.

\subsection{Ablation Studies}
We conduct ablation studies using the Qwen3-0.6B model to evaluate the contribution of \( p_{\text{align}} \) and \( p_{\text{cheat}} \) in the reward function. Specifically, we remove \( p_{\text{align}} \) (retaining \( p_{\text{cheat}} \)) and remove \( p_{\text{cheat}} \) (retaining \( p_{\text{align}} \)) from the reward function, respectively. The experimental results are shown in Table \ref{tab:ablation}, ablating any single reward term leads to performance degradation. Notably, removing \( p_{\text{align}} \) results in a more significant drop, indicating that \( p_{\text{align}} \) plays a critical role in aligning the data distribution; without it, model performance remains constrained due to the mismatch in data distribution. Meanwhile, removing \( p_{\text{cheat}} \) also causes performance loss, confirming that this anti-cheating term effectively prevents the model from exploiting shortcuts for cheating that would lead to final performance degradation, thereby providing an important constraint for DART. These findings fully demonstrate that both \( p_{\text{align}} \) and \( p_{\text{cheat}} \) play necessary and complementary roles in DART.

\begin{table}[t]
\centering
\caption{Ablation study of DART on six datasets: SV (SVAMP), BE (Boolean Expressions), WoL (Web of Lies), PQA (ProntoQA), MB (MBPP), HE (Human-Eval), and Avg. (average). The \textbf{bold} indicates the best results.}
\label{tab:ablation}
\small
\setlength{\tabcolsep}{3.2pt}
\renewcommand{\arraystretch}{1.08}
\begin{tabular}{l c | ccc | cc | c}
\toprule
\textbf{Variant} & \textbf{SV} & \textbf{BE} & \textbf{WoL} & \textbf{PQA} & \textbf{MB} & \textbf{HE} & \textbf{Avg.} \\
\midrule
DART & \textbf{81.4} & \textbf{73.9} & 59.2 & \textbf{63.1} & 34.9 & \textbf{38.1} & \textbf{58.4} \\
w/o $P_{\mathrm{align}}$ & 81.1 & 70.5 & 60.8 & 60.3 & \textbf{35.1} & 33.5 & 56.9 \\
w/o $P_{\mathrm{cheat}}$ & 80.6 & 73.4 & \textbf{64.3} & 61.8 & 33.7 & 32.3 & 57.7 \\
\bottomrule
\end{tabular}
\end{table}

\subsection{Performance Across Different Reward Functions} 
To evaluate the rationality of our reward-function design, we investigate several alternative reward functions. The most instructive approaches based on Qwen3-0.6B are presented below, as shown in Table \ref{tab:qwen3-reward}:
(1) The threshold \( \tau \) is set to 50\%, with an upper bound on the penalty. When the model produces an incorrect answer, the reward is set to a minimum value of 0; otherwise, the maximum penalty is 0.95. Although this version of the reward function is relatively simple, resulting in limited effectiveness in data optimization, it still demonstrates the potential of our method. (2) The threshold \( \tau \) is set to 35\%, and an exponential smoothing approach is adopted for scoring, which maps the reward value to a range between 0 and 1. This allows for a more intuitive observation of reward dynamics and convergence, making the score more stable. A cheating prevention mechanism is also incorporated to prevent the model from obtaining scores by directly outputting answers.

\begin{table*}[!ht]
\centering
\caption{Performance (\%) of Qwen3-0.6B model across different reward functions. The \textbf{bold} and \underline{underline} indicate the best and second-best results, respectively.}
\small
\renewcommand{\arraystretch}{1.1}
\setlength{\tabcolsep}{3pt}
\begin{tabular*}{\textwidth}{@{\extracolsep{\fill}}l|c|ccc|cc|c}
\toprule
& \multicolumn{1}{c|}{\textbf{Mathematical}} & \multicolumn{3}{c|}{\textbf{Logical Reasoning}} & \multicolumn{2}{c|}{\textbf{Code Capability}} 
& \\
\textbf{Method} & \textbf{SVAMP} & \textbf{Boolean Expressions} & \textbf{Web of Lies} & \textbf{ProntoQA} & \textbf{MBPP} & \textbf{Human-Eval} & \textbf{Average} \\
\midrule
\multicolumn{8}{c}{\textit{Models based on Qwen3-0.6B}} \\
\midrule
\textbf{DART (ours)} & 81.37 & \textbf{73.90} & 59.20 & \textbf{63.05} & \textbf{34.91} & \underline{38.11} & \textbf{58.42} \\
Scheme 1  & \underline{81.60} & 70.00 & \textbf{60.90} & \underline{59.65} & 34.42 & 37.04 & 57.27 \\
Scheme 2  & \textbf{81.97} & \underline{72.80} & \underline{60.10} & 59.30 & \underline{34.65} & \textbf{38.26} & \underline{57.85} \\
\bottomrule
\end{tabular*}
\label{tab:qwen3-reward}
\end{table*}

\begin{table*}[!ht]
\centering
\caption{Performance (\%) using the mapper model based on Qwen3-0.6B for ARC-Challenge CoT optimization. The \textbf{bold} and \underline{underline} indicate the best and second-best results, respectively.}
\small
\renewcommand{\arraystretch}{1.1}
\setlength{\tabcolsep}{3pt}
\begin{tabular*}{\textwidth}{@{\extracolsep{\fill}}l|c|ccc|cc|c}
\toprule
& \multicolumn{1}{c|}{\textbf{Mathematical}} & \multicolumn{3}{c|}{\textbf{Logical Reasoning}} & \multicolumn{2}{c|}{\textbf{Code Capability}} 
& \\
\textbf{Method} & \textbf{SVAMP} & \textbf{Boolean Expressions} & \textbf{Web of Lies} & \textbf{ProntoQA} & \textbf{MBPP} & \textbf{Human-Eval} & \textbf{Average} \\
\midrule
\multicolumn{8}{c}{\textit{Models based on Qwen3-0.6B}} \\
\midrule
Base & 79.03 & 72.60 & \underline{50.50} & 59.95 & 34.01 & \underline{37.65} & 55.62 \\
Original-SFT & 73.92 & 68.00 & 25.30 & \underline{62.85} & 33.70 & 33.99 & 49.63 \\
GRPO & \underline{80.00} & \textbf{73.70} & 49.50 & 60.65 & \underline{34.52} & 37.50 & \underline{55.98} \\
\textbf{DART(ours)} & \textbf{80.05} & \underline{73.10} & \textbf{59.30} & \textbf{65.30} & \textbf{35.19} & \textbf{41.62} & \textbf{59.09} \\
\bottomrule
\end{tabular*}
\label{tab:qwen3-transfer}
\end{table*}

\subsection{Transfer Experiment} To further verify the applicability and stability of our method, we design a transfer experiment for CoT translation across different datasets. In addition to MATH\cite{DBLP:journals/corr/abs-2103-03874} used during mapper training, ARC-Challenge\cite{clark2018thinksolvedquestionanswering} is selected as the analysis target. By employing the pre-trained mapper model based on Qwen3-0.6B from our experiments, we translate and reconstruct the original CoT of ARC-Challenge. This process aims to verify whether the mapper model has learned a general normative framework for logical expression that transcends specific subject domains, and whether it can transform the original reasoning paths into reasoning trajectories that better align with the model's distribution. As shown in Table \ref{tab:qwen3-transfer}, the results demonstrate that our method exhibits transferability, enabling improvements in model performance across different training datasets.

\subsection{Parameter Analysis}
\begin{figure}[t]
    \centering
     \includegraphics[width=\linewidth]{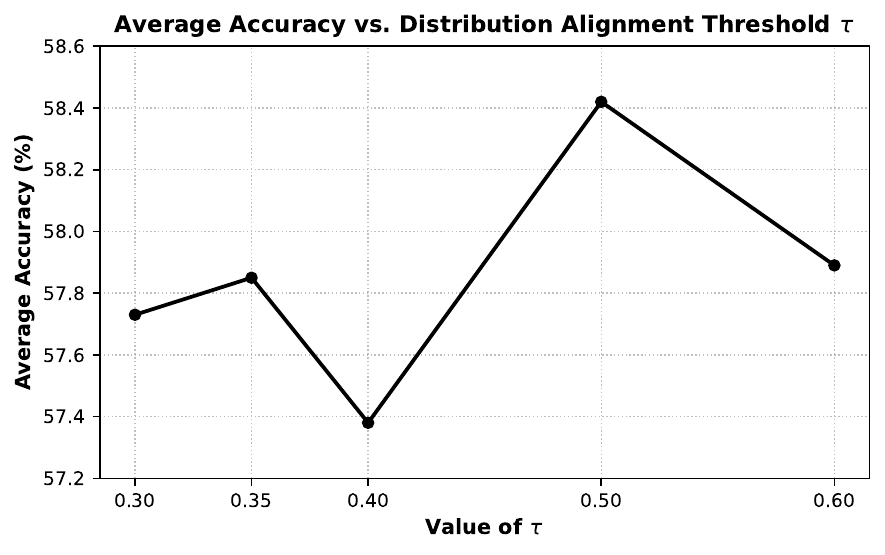}
    \caption{Average accuracy of Qwen3-0.6B across six datasets under different alignment thresholds $\tau$.}
    \label{fig:tau_analysis}
\end{figure}

To investigate the impact of the distribution alignment threshold $\tau$ in the reward function, we conduct a systematic analysis on Qwen3-0.6B, with results shown in Figure~\ref{fig:tau_analysis}. When \( \tau \) is too small, the distribution constraint becomes overly loose, making the model more prone to deviating from the target data distribution. As $\tau$ increases, performance gradually improves and reaches the best result at $\tau=0.5$. However, further increasing \( \tau \) leads to performance degradation, which may be attributed to the overly strict constraint incorrectly penalizing effective exploration behaviors. In addition, an excessively large threshold may produce sparse reward signals and reduce training stability. Overall, the model remains relatively robust across a broad range of $\tau$ values, with $\tau=0.5$ achieving the best balance between performance and stability.

\section{Conclusion}
In this work, we investigate self-optimization of SFT data under distribution mismatch with the target model, without relying on external teacher models. We formulate this problem as optimizing demonstration transformations to better align training data with the target model distribution. Based on this formulation, we propose DART, a general RL-based data optimization framework that learns a mapper model, rewrites the original demonstrations, and fine-tunes the target model on the optimized data. Extensive experiments show that DART consistently improves generalization and reasoning performance across diverse reasoning tasks, outperforming strong baselines while demonstrating robust effectiveness.

\section*{Limitations}
Acknowledging the limitations of this study is also important. Due to hardware constraints, we were unable to conduct experiments on larger models (e.g., those exceeding 3B parameters). Additionally, the design of the reward function during RL training contains certain flaws, as it does not fully eliminate the possibility of the model adopting other extreme strategies to cheat. Nevertheless, the proposed data optimization method offers a new perspective for enhancing the reasoning capabilities of language models and remains promising for scaling to larger models, as the RL paradigm can enhance exploration efficiency and reasoning robustness. The core strength of our method lies in constructing datasets that align with the actual data distribution of the model, which brings about favorable scalability and generalization ability.

\section*{Ethical Statements}
This study does not raise any ethical concerns. All data and models used are publicly available, and any sensitive information has been redacted.

\section*{Acknowledgments}
This work was supported by the National Key Research and Development Program of China (Grant No. 2025YFF1505704), the National Science and Technology Major Project (Grant No. 2022ZD0117402), the National Natural Science Foundation of China (Grant No. 62441617), and the Beijing Advanced Innovation Center for Future Blockchain and Privacy Computing.


\bibliography{custom}

\appendix

\section{Proof of Theorem 1}
\label{appendix:proof}
For a reasoning pattern $z$, the effective SFT supervision probability under any data distribution $p$ is
\[
\rho_p(z) = P_p(\mathcal{C}_z \cap \mathcal{M}_z \mid z).
\]
By the chain rule of conditional probability, we have
\[
\rho_p(z) = P_p(\mathcal{C}_z \mid z) P_p(\mathcal{M}_z \mid \mathcal{C}_z,z).
\]
For Original-SFT, this gives
\[
\begin{aligned}
\rho_{p_0}(z)
&= P_{p_0}(\mathcal{C}_z \mid z)
   P_{p_0}(\mathcal{M}_z \mid \mathcal{C}_z,z) \\
&= c_0(z)m_0(z).
\end{aligned}
\]
Since the optimized samples have passed correctness
verification, we have
\[
P_{p_\phi}(\mathcal{C}_z \mid z)=1.
\]
By the distribution matching assumption,
\[
P_{p_\phi}(\mathcal{M}_z \mid \mathcal{C}_z,z) \ge m_0(z)+\Delta_m(z).
\]
Therefore,
\[
\begin{aligned}
\rho_{p_\phi}(z)
&= P_{p_\phi}(\mathcal{C}_z \mid z)
   P_{p_\phi}(\mathcal{M}_z \mid \mathcal{C}_z,z) \\
&\ge m_0(z)+\Delta_m(z).
\end{aligned}
\]
Since $0\le c_0(z)\le 1$, we have
\[
\rho_{p_0}(z) = c_0(z)m_0(z) \le m_0(z).
\]
Because $\Delta_m(z)>0$,
\[
m_0(z)+\Delta_m(z) > m_0(z) \ge c_0(z)m_0(z).
\]
Thus,
\[
\rho_{p_\phi}(z) > \rho_{p_0}(z).
\]
Next, let $N_p(z)$ be the number of effective supervision samples for pattern $z$ among $n_z$ training samples. Then
\[
N_p(z) \sim \mathrm{Binomial}(n_z,\rho_p(z)).
\]
The probability that the model observes at least $k_z$ effective samples is
\[
\begin{aligned}
\Psi_{n_z,k_z}(\rho_p(z))
&= P(N_p(z)\ge k_z) \\
&= \sum_{j=k_z}^{n_z} 
\binom{n_z}{j}
\rho_p(z)^j
\bigl(1-\rho_p(z)\bigr)^{n_z-j}.
\end{aligned}
\]
For fixed $n_z$ and $k_z$, this binomial tail probability is monotonically increasing in $\rho_p(z)$. Since
\[
\rho_{p_\phi}(z) > \rho_{p_0}(z),
\]
we have
\[
\Psi_{n_z,k_z}(\rho_{p_\phi}(z)) > \Psi_{n_z,k_z}(\rho_{p_0}(z)).
\]
Let $a_-(z)$ denote the expected accuracy when the reasoning pattern $z$ is not successfully acquired, and let $a_+(z)$ denote the expected accuracy after successful acquisition, with
\[
a_+(z)>a_-(z).
\]
Then the expected post-SFT accuracy under distribution $p$ can be written as
\[
\mathrm{Acc}_p(z) = a_-(z) + \bigl(a_+(z)-a_-(z)\bigr) \Psi_{n_z,k_z}(\rho_p(z)).
\]
Therefore,
\[
\begin{aligned}
\mathrm{Acc}_{p_\phi}(z)-\mathrm{Acc}_{p_0}(z)
&= \bigl(a_+(z)-a_-(z)\bigr) \\
&\quad \cdot
\Bigl[
\Psi_{n_z,k_z}(\rho_{p_\phi}(z)) \\
&\qquad\quad
-\Psi_{n_z,k_z}(\rho_{p_0}(z))
\Bigr].
\end{aligned}
\]
Since both factors on the right-hand side are positive, we obtain
\[
\mathrm{Acc}_{p_\phi}(z) > \mathrm{Acc}_{p_0}(z).
\]
Taking expectation over test reasoning patterns
$z\sim p_{\mathrm{test}}(z)$ yields
\[
\mathrm{Acc}_{\mathrm{Optimized}} > \mathrm{Acc}_{\mathrm{Original\text{-}SFT}}.
\]
This completes the proof.

\section{Experimental Details}
\label{appendix:exp_details}
\subsection{Datasets for Main Experiments}
\label{appendix:dataset_details}
We summarize the number of samples in the training and evaluation datasets in Table \ref{tab:datasets_statics}. The MATH\cite{DBLP:journals/corr/abs-2103-03874} dataset is used as the primary training set, from which samples of Levels 3 to 5 are selected for training, and the original CoT is generated by DeepSeek-V3\cite{deepseekai2025deepseekv3technicalreport}. ARC-Challenge\cite{clark2018thinksolvedquestionanswering} is adopted for the transfer experiment. All other datasets are used exclusively for evaluation.

\begin{table}[t]
\centering
\caption{Statistics of the training and test datasets used by all methods.}
\label{tab:datasets_statics}
\small
\setlength{\tabcolsep}{4pt}
\renewcommand{\arraystretch}{1.1}
\begin{tabular}{lll}
\toprule
\textbf{Dataset} & \textbf{Description
} & \textbf{\# Train / Test} \\
\midrule
MATH Level3-5            & Math Reasoning     & 5000 / - \\
SVAMP   & Math Reasoning     & - / 1000 \\
Boolean Expressions   & Logical Reasoning     & - / 250 \\
Web of Lies            & Logical Reasoning     & - / 250 \\
ProntoQA            & Logical Reasoning     & - / 500 \\
MBPP          & Code     & - / 974 \\
Human-Eval           & Code     & - / 164 \\
ARC-Challenge           & Scientific Reasoning     & 2492 / - \\
\bottomrule
\end{tabular}
\end{table}

\subsection{Training and Comparison Details}
\label{appendix:imple_details}
All experiments are implemented using the VERL framework \cite{Sheng_2025} with Python 3.10 and PyTorch 2.6. We generate original CoT reasoning traces on the MATH training dataset using DeepSeek-V3\cite{deepseekai2025deepseekv3technicalreport}. These traces are subsequently translated and optimized in Stage II of our method for use in subsequent SFT.
In the reward function design, we compute \( P_{\text{cheat}} \) based on length.
\[
P_{\text{cheat}} =
\begin{cases} 
1.0, & L_{\text{valid}} \geq L_{\text{min}} \\
0.0, & L_{\text{valid}} < L_{\text{min}}
\end{cases}
\]
During the RL training phase, the model is trained for 1 epoch with a global batch size of 32. The maximum response length is set to 4096, the learning rate is $1 \times 10^{-6}$, and the KL divergence loss coefficient is 0.001. The maximum number of sequences processed simultaneously is 256, with 8 responses sampled per prompt. The mini-batch size for each PPO update is 16.
In the SFT phase, the maximum sequence length is 4096, the model is trained for 3 epochs, and the learning rate is $2 \times 10^{-5}$.
During the evaluation phase, the temperature is set to 0.7, top-p is 0.9, the maximum response length is 4096, the random seed is set to 42, and 4 responses are generated per prompt. For GRPO, we adopt the same parameter settings as DART. For STaR, we sample 8 responses per question and only retain the correct ones for SFT. For TESSY, we use DeepSeek-V3 and base models to alternately generate answers, producing the dataset for SFT.

\subsection{Comparison of Computational Cost}
To compare the training overhead of all methods, we report the total runtime measured on a single NVIDIA A100 GPU with 40GB memory, using Qwen3-0.6B as the base model, as summarized in Table \ref{tab:time_cost}. During the construction phase of the original CoT, taking DeepSeek-V3 as an example, due to the relatively slow API responses, we employed 20 concurrent terminals but still required 8 hours to complete the dataset construction, in addition to incurring extra API usage costs. Data processing time for DART and Original-SFT already includes the cost of generating the original CoT. For Original-SFT, the SFT process takes 0.4 hours. For GRPO, training converges within 7 hours. For TESSY, data processing takes 12 hours and SFT takes 0.5 hours. For STaR, data processing takes 2 hours and SFT takes 0.5 hours.
For DART, Step I trains the mapper model to convergence in 6 hours. Step II performs CoT translation optimization and filtering, taking 0.5 hours. Step III conducts SFT, taking 0.4 hours, resulting in a total runtime of 6.9 hours. Notably, although the mapper training in DART takes 6 hours, once trained, the mapper model can be reused for CoT translation and optimization across different datasets or different base models, with minimal subsequent SFT cost. These results highlight that DART improves exploration efficiency and convergence speed by aligning the data distribution. This offers a new perspective for enhancing the reasoning capability and robustness of LLMs.

\begin{table}[t]
\centering
\caption{Estimated computational time (hours) for various methods with Qwen3-0.6B.}
\label{tab:time_cost}
\small
\setlength{\tabcolsep}{4pt}
\renewcommand{\arraystretch}{1.1}
\begin{tabular}{lcccc}
\toprule
\textbf{Method} & \textbf{Data Processing} & \textbf{Training} & \textbf{SFT} & \textbf{Total} \\
\midrule
DART        & 8 & 6.5 & 0.4 & 13.9 \\
Original-SFT       & 8 & 0 & 0.4 & 8.4 \\
GRPO       & 0 & 7 & 0.5 & 7.5 \\
TESSY       & 12 & 0 & 0.5 & 12.5 \\
STaR       & 2 & 0 & 0.5 & 2.5 \\
\bottomrule
\end{tabular}
\end{table}

\section{Pseudocode of DART}
\label{pseudocode}
We present the pseudocode in Algorithm \ref{alg:dart}.
\begin{algorithm}[t]
\caption{DART: Data Adaptation for Reasoning Tuning}
\label{alg:dart}
\begin{algorithmic}
\REQUIRE
    Dataset $\mathcal{D} = \{(x_i, y_i)\}_{i=1}^{n}$,
    Base model $\pi_{\theta}$, Mapper model $M_{\phi}$(initialized from $\pi_{\theta}$),
    Reference model $\pi_{\text{ref}}$ (initialized from $\pi_{\theta}$)
    
\FOR{epoch $= 1$ to $E_{\text{RL}}$}

    \STATE \textbf{Stage I: Mapper Training via Reinforcement Learning}

    \FOR{each sample $(x_i, y_i) \in \mathcal{D}$}
        \STATE Generate transformed CoT: $\tilde{y}_i \sim M_{\phi}(\cdot \mid x_i, y_i)$
        \STATE Compute answer correctness reward $R_{\text{base}}$
        \STATE Compute distribution alignment penalty $P_{\text{align}}$ using reference model $\pi_{\text{ref}}$
        \STATE Compute anti-cheating penalty $P_{\text{cheat}}$
        \STATE Compute final reward: $R_{\text{final}} = R_{\text{base}} \cdot P_{\text{align}} \cdot P_{\text{cheat}}$
        \STATE Update mapper $M_{\phi}$ using GRPO with reward $R_{\text{final}}$
    \ENDFOR

    \STATE \textbf{Stage II: Optimized Data Synthesis}

    \STATE Initialize empty dataset $\tilde{\mathcal{D}} = \emptyset$
    \FOR{each sample $(x_i, y_i) \in \mathcal{D}$}
        \STATE Generate transformed CoT: $\tilde{y}_i \sim M_{\phi}(\cdot \mid x_i, y_i)$
        \IF{correctness verification and logical density analysis pass}
            \STATE Add $(x_i, \tilde{y}_i)$ to $\tilde{\mathcal{D}}$
        \ENDIF
    \ENDFOR

    \STATE \textbf{Stage III: Supervised Fine-Tuning}
    
    \STATE Initialize base model $\pi_{\theta}$ from pre-trained weights
    \FOR{epoch $= 1$ to $E_{\text{SFT}}$}
        \FOR{each mini-batch $\mathcal{B} \subset \tilde{\mathcal{D}}$}
            \STATE Compute SFT loss:
            $\mathcal{L}_{\text{SFT}}(\theta)
            = - \frac{1}{|\mathcal{B}|}
            \sum_{(x_i,\tilde{y}_i)\in \mathcal{B}}
            \log \pi_{\theta}(\tilde{y}_i \mid x_i)$
            \STATE Update $\theta$ by gradient descent on $\mathcal{L}_{\text{SFT}}$
        \ENDFOR
    \ENDFOR

\ENDFOR

\STATE \textbf{Return} fine-tuned model $\pi_{\theta^*}$

\end{algorithmic}
\end{algorithm}

\section{Use of Large Language Models}
Some parts of the text were refined with the assistance of LLMs. All content and responsibility for the work remain with the authors.

\section{Prompts}
\label{appendix:prompt_details}
During the experiments, prompts are required for all model usages. The prompt used for generating the original CoT is shown in Figure \ref{fig:originalcot_prompt}, the prompt for training the mapper model is shown in Figure \ref{fig:train_prompt}, and the prompt for the evaluation phase is shown in Figure \ref{fig:eval_prompt}.

\section{Case Study}
\label{appendix:case_study}
To further demonstrate the optimization effect of the mapper model \(M\) on reasoning trajectories, we present a representative geometry reasoning example as a case study in Figure \ref{fig:case_geometry}. Although the original CoT is able to derive the correct answer, its reasoning process is relatively verbose, involving extensive intermediate steps such as coordinate construction, vector computation, quadratic equation solving, and sign verification. Such detail-intensive reasoning trajectories that deviate from the target data distribution may increase the difficulty for small-scale models to learn and reproduce effective reasoning patterns during training.
After being transformed by the mapper model \(M\), the optimized CoT preserves the core geometric relations and area computation logic of the original solution while removing a large number of redundant coordinate derivations, vector expansions, and intermediate verification steps. Compared with the original CoT, the transformed reasoning process becomes significantly more concise and structured, making the overall reasoning style better aligned with the learning distribution of small models.
This case demonstrates that the mapper model does not merely compress the response length. Instead, it restructures and simplifies the original reasoning trajectory while preserving both correctness and the essential reasoning structure. Such transformations can effectively reduce redundant reasoning information, make the supervision signal more concentrated, and improve the compatibility between reasoning trajectories and the target model distribution, thereby enabling small models to learn more stable and effective reasoning patterns.
\begin{figure*}[!t]
\begin{tcolorbox}[
    title=Instance-Level Comparison of Original and Optimized CoT ,
    colback=gray!5,
    colbacktitle=black!80,
    coltitle=white,
    colframe=black!70,
    arc=5pt,
    fonttitle=\bfseries,
    width=\textwidth
]
\small
\textbf{Problem:}
\begin{verbatim}
In triangle $ABC$, $\tan \angle CAB = \frac{22}{7}$, and the altitude
from $A$ divides $\overline{BC}$ into segments of length $3$ and $17$.
What is the area of triangle $ABC$? Let's think step by step and output
the final answer within \boxed{}.
\end{verbatim}
\textbf{Original CoT:}
\begin{verbatim}
Alright, let's go step-by-step.
Let D be the foot of the altitude from A to BC. Place D=(0,0),
B=(-3,0), C=(17,0), and A=(0,h).
Then \overrightarrow{AB}=(-3,-h), and \overrightarrow{AC}=(17,-h).
Using the tangent formula between two vectors,
\tan \angle CAB =
|\overrightarrow{AB} \times \overrightarrow{AC}|
/
(\overrightarrow{AB} \cdot \overrightarrow{AC})
= 20h / (h^2 - 51).
Since \tan \angle CAB = 22/7,
20h / (h^2 - 51) = 22/7.
Cross-multiplying gives
140h = 22(h^2 - 51).
Thus,
11h^2 - 70h - 561 = 0.
Using the quadratic formula,
h = (70 \pm \sqrt{70^2 + 4 \cdot 11 \cdot 561}) / 22
  = (70 \pm 172) / 22.
Hence, h = 11.
We also verify that h^2 - 51 = 70 > 0, which is consistent
with the sign condition.
Finally, BC = 3 + 17 = 20, so the area is
1/2 * 20 * 11 = 110.
Thus, the final answer is \boxed{110}.
\end{verbatim}
\textbf{Optimized CoT by Mapper:}
\begin{verbatim}
To find the area of triangle $ ABC $, we use the given information:
- $ \tan \angle CAB = \frac{22}{7} $
- The altitude from $ A $ divides $ BC $ into segments of length 3 and 17,
  so $ BC = 3 + 17 = 20 $
- Let $ h $ be the height from $ A $ to $ BC $, so $ A = (0, h) $
Using the formula for the tangent of the angle between two vectors
$ \overrightarrow{AB} $ and $ \overrightarrow{AC} $, we find that $ h = 11 $.
The area of triangle $ ABC $ is then:
$$
\text{Area} = \frac{1}{2} \times \text{base} \times \text{height}
= \frac{1}{2} \times 20 \times 11 = 110
$$
**Final answer:**
$$
\boxed{110}
$$
\end{verbatim}
\end{tcolorbox}
\caption{Illustrative comparison between the original CoT and the Optimized CoT generated by the mapper model based on Qwen3-0.6B.}
\label{fig:case_geometry}
\end{figure*}

\clearpage
\begin{figure*}[!t]
\begin{tcolorbox}[
    title=Prompt for Original CoT Generation ,
    colback=gray!5,
    colbacktitle=black!80,
    coltitle=white,
    colframe=black!70,
    arc=5pt,
    fonttitle=\bfseries,
    width=\textwidth
]
You are an expert assistant teacher specializing in math and reasoning. 
For each task, first provide a detailed step-by-step reasoning process and then give the final answer
enclosed in \texttt{\textbackslash boxed\{\}}.
\end{tcolorbox}
\caption{Prompt used for generating the original CoT}
\label{fig:originalcot_prompt}
\end{figure*}

\begin{figure*}[!t]
\begin{tcolorbox}[
    title=Mapper Training Prompt,
    colback=gray!5,
    colbacktitle=black!80,
    coltitle=white,
    colframe=black!70,
    arc=5pt,
    fonttitle=\bfseries,
    width=\textwidth
]
Below is a question and a reference answer. Please restate the answer in your own words based on your understanding. You may change the wording, remove unnecessary steps, and express it in a simpler and clearer way that you can easily understand. Directly output your final answer. Question: \{QUESTION\}. Let's think step by step and output the final answer within \texttt{\textbackslash boxed\{\}}. Reference Answer: \{REFERENCE ANSWER\}
\end{tcolorbox}
\caption{Prompt used for training the Mapper model \( M \)}
\label{fig:train_prompt}
\end{figure*}

\begin{figure*}[!t]
\begin{tcolorbox}[
    title=Evaluate Prompt,
    colback=gray!5,
    colbacktitle=black!80,
    coltitle=white,
    colframe=black!70,
    arc=5pt,
    fonttitle=\bfseries,
    width=\textwidth
]
For all evaluation datasets: \\
Question: \{QUESTION\}. Let's think step by step and output the final answer within \texttt{\textbackslash boxed\{\}}. \\
assistant \\
\end{tcolorbox}
\caption{Prompt used for evaluating all models.}
\label{fig:eval_prompt}
\end{figure*}

\end{document}